\documentclass[runningheads]{llncs}
\usepackage[noend]{algpseudocode}
\usepackage{algorithmicx,algorithm}
\usepackage{graphicx}
\usepackage{epstopdf}
\usepackage{bm}
\usepackage{eucal}
\usepackage{amsmath}
\usepackage{amsfonts,amssymb}
\usepackage{soul}
\usepackage{url}
\usepackage[hidelinks]{hyperref}
\usepackage[small]{caption}
\usepackage{booktabs}
\usepackage{enumerate}
\usepackage{soul}
\usepackage{color}
\usepackage{comment}
\usepackage{subfigure}
\usepackage{multirow}
\begin{document}
\title{{Partial Multi-label Learning with Label and Feature Collaboration}
\thanks{This work is supported by NSFC (61872300 and 61871010). Corresponding author: gxyu@swu.edu.cn (Guoxian Yu).}}
\author{Tingting Yu\inst{1}, Guoxian Yu\inst{1},  Jun Wang\inst{1}, Maozu Guo\inst{2}}
\authorrunning{T. Yu, G. Yu, J. Wang and M. Guo}
\institute{College of Comp. \& Inf. Sci., Southwest University, Chongqing, China
\and College of Elec. \& Inf. Eng., Beijing Univ. of Civil Eng. and Arch., Beijing, China\\
\email{\{ttyu, gxyu, kingjun\}@swu.edu.cn,  maozuguo@becea.edu.cn}}
%
\maketitle              
\begin{abstract}
Partial multi-label learning (PML) models the scenario where each training instance is annotated with a set of candidate labels, and only some of the labels are relevant. The PML problem is practical in real-world scenarios, as it is difficult and even impossible to obtain precisely labeled samples. Several PML solutions have been proposed to combat with the prone misled by the irrelevant labels concealed in the candidate labels, but they generally focus on the smoothness assumption in feature space or low-rank assumption in label space, while ignore the negative information between features and labels. Specifically, if two instances have largely overlapped candidate labels, irrespective of their feature similarity, their ground-truth labels should be similar; while if they are dissimilar in the feature and candidate label space, their ground-truth labels should be dissimilar with each other. To achieve a credible predictor on PML data, we propose a novel approach called PML-LFC (Partial Multi-label Learning with Label and Feature Collaboration). PML-LFC estimates the confidence values of relevant labels for each instance using the similarity from both the label and feature spaces, and trains the desired predictor with the estimated confidence values.  PML-LFC achieves the predictor and the latent label matrix in a reciprocal reinforce manner by a unified model, and develops an alternative optimization procedure to optimize them. Extensive empirical study on both synthetic and real-world datasets demonstrates the superiority of PML-LFC.

\end{abstract}

\section{Introduction}
Multi-label learning (MLL) deals with the scenario where each instance is annotated with a set of discrete non-exclusive labels \cite{zhang2014mllreview,gibaja2015mllsurvey}. Recent years have witnessed an increasing research and application of MLL in various domains, such as image annotation~\cite{wu2018IJCV}, cybersecurity~\cite{han2018kdd}, gene functional annotation~\cite{yu2016SimNet}, and so on. Most MLL methods have an implicit assumption that each training example is precisely annotated with all of its relevant labels. However, it is difficult and costly to obtain fully annotated training examples in most real-word MLL tasks \cite{tu2018mljmf}. Therefore, recent MLL methods not only focus on how to assign a set of appropriate labels to unlabeled examples using the label correlations ~\cite{zhang2010kdd,zhu2017tkde}, but also on replenishing missing labels for incompletely labeled samples~\cite{sun2010well,wu2018IJCV,tan2018IJCAI}.

\begin{figure}[h!t]
\centering
\includegraphics[width=9cm, height=4cm]{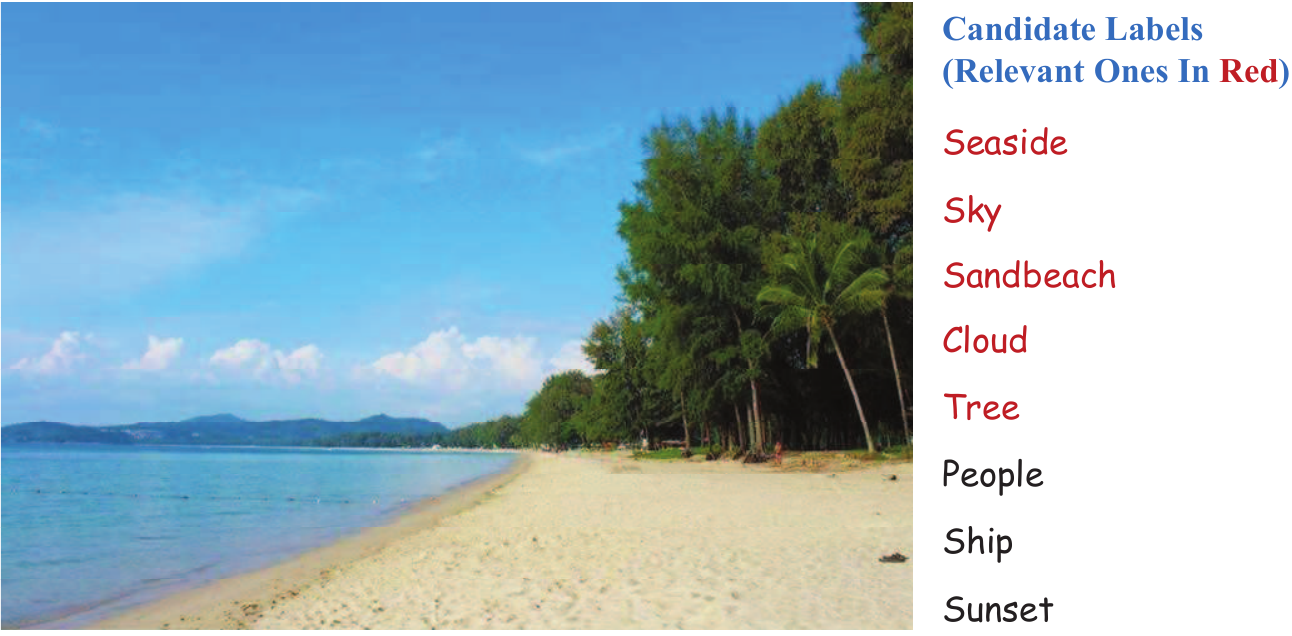}
\caption{An exemplary partial multi-label learning scenario. The image is annotated with eight candidate labels, only the first five (in {\color{red}red}) are relevant, and the last three (in black) are irrelevant.}
\label{figure1}
\end{figure}

Existing MLL solutions still overlook another fact that naturally arises in real-world scenarios. For example, in Fig.\ref{figure1}, the image was crowdly-annotated by  workers with  `Seaside', `Sky', `Sandbeach', `Cloud', `Tree', `People', `Sunset' and `Ship'. Among these labels, the first five are relevant, and the last three are irrelevant of this image.  Obviously, the training procedure is prone to be misled by irrelevant labels concealed in the candidate labels of training samples. To combat with such major difficulty, some pioneers term learning on such training data with irrelevant labels as \emph{Partial Multi-label Learning} (PML) \cite{xie2018PML,yu2018fPML}, and proposed several PML approaches  \cite{wang2019DRAMA,fang2019PRACTICLE,sun2019PML} to identify the irrelevant labels concealed in the candidate labels of annotated samples, and to achieve a predictor robust (or less prone) to irrelevant labels of training data.

\begin{figure*}[h!t]
\centering
\hspace{-2em}
\subfigure[Two instances with both high feature and semantic similarity.]{\label{figure2a}\includegraphics[width=5.5cm, height=3cm]{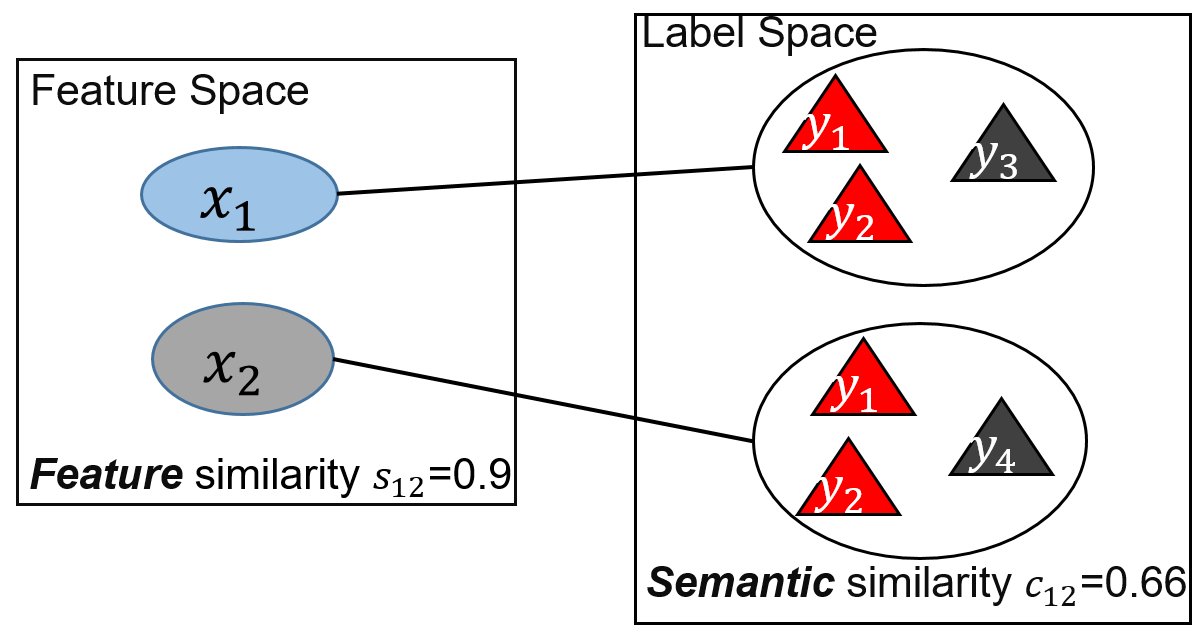}}
\hspace{1em}
\subfigure[Two instances with high feature similarity but with low (zero) semantic similarity.]{\label{figure2b}\includegraphics[width=5.5cm, height=3cm]{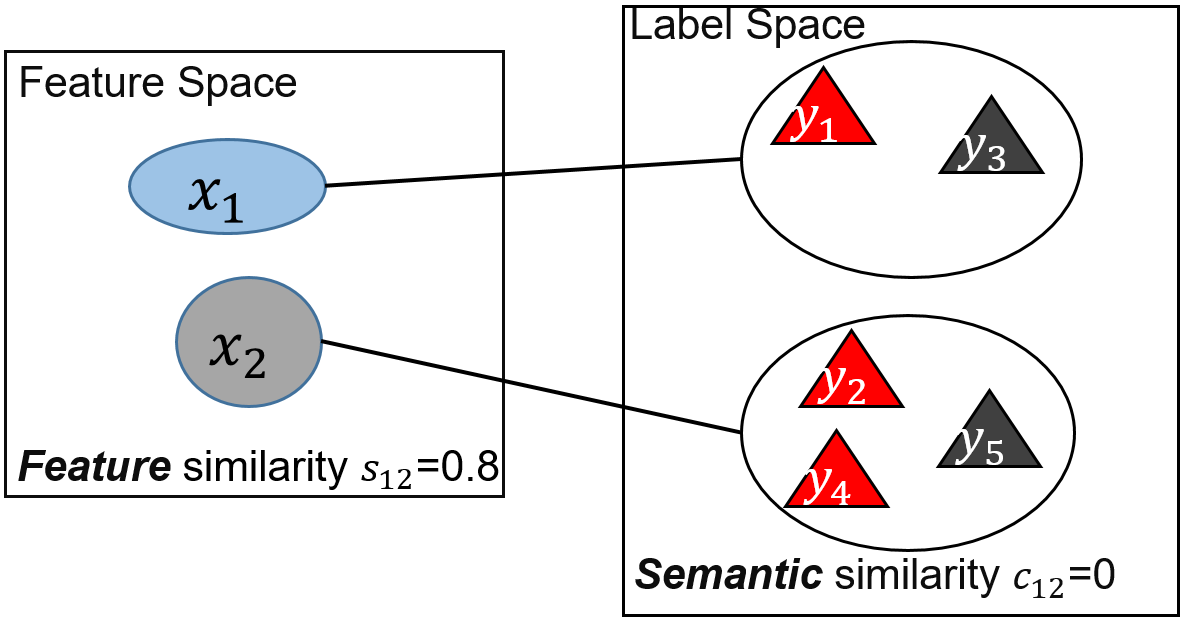}}
\vspace{-2em}
\hspace{-2em}
\subfigure[Two instances with low (moderate) feature similarity but with high semantic similarity.]{\label{figure2c}\includegraphics[width=5.5cm, height=3cm]{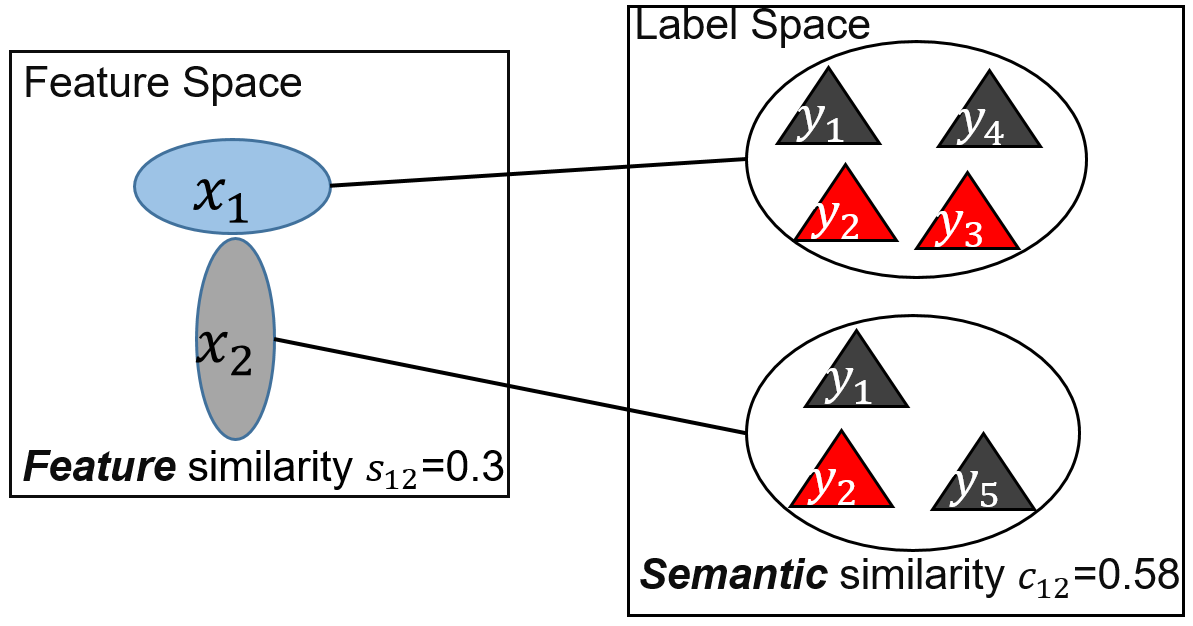}}
\hspace{1em}
\subfigure[Two instances with low feature similarity and low semantic similarity.]{\label{figure2d}\includegraphics[width=5.5cm, height=3cm]{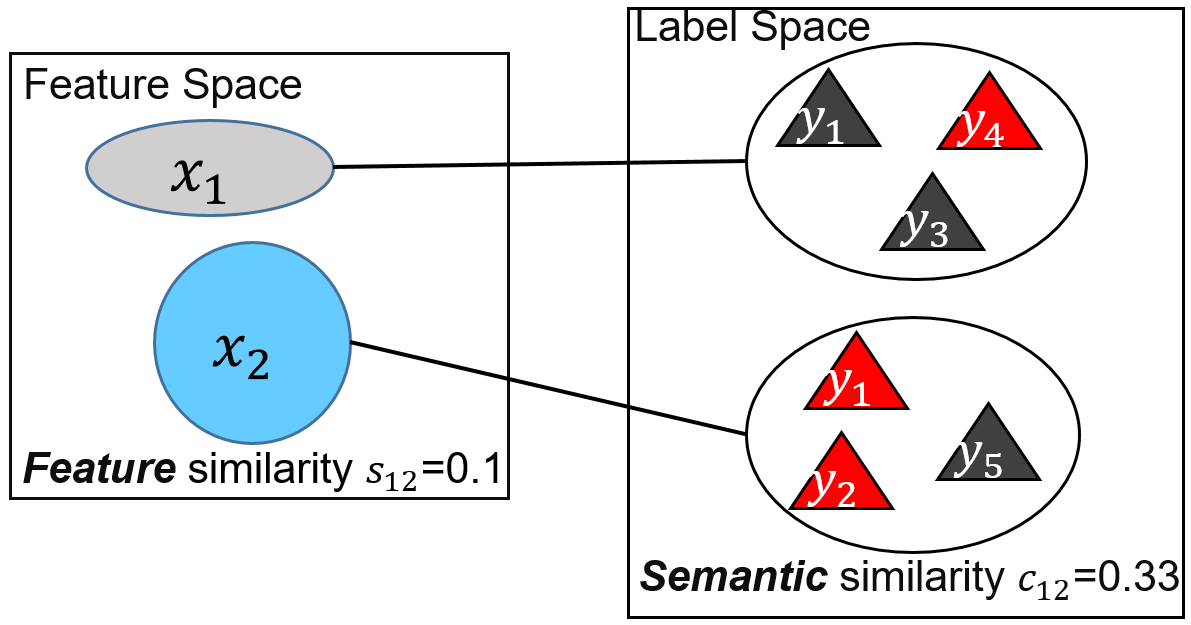}}
\caption{The feature and label information of PML data in four scenarios. The semantic similarity is computed on the candidate labels of instances, the feature similarity is computed on the feature vectors of instances. Ground-truth labels are highlighted in red, while other candidate labels are in grey.}
\label{figure2}
\end{figure*}

However, contemporary approaches either mainly focus on the smoothness assumption that the (dis-)similar instances should have (dis-)similar ground-truth labels~\cite{xie2018PML,wang2019DRAMA,fang2019PRACTICLE}, or the low-rank assumption that the ground-truth label matrix should be low-rank \cite{yu2018fPML,sun2019PML}. While these two assumptions can not well handle the case that two instances without any common candidate label but with high feature similarity, and two instances with some overlapped candidate labels but with a low feature similarity. As a result, the smoothness-based methods~\cite{xie2018PML,wang2019DRAMA,fang2019PRACTICLE} ignore the \emph{negative} information  that two instances with high (low) feature similarity but with a low (high) semantic similarity from the candidate labels.  In other words, these methods do not make a good collaborative use of the information from the label and feature space for PML. For example, in Fig.\ref{figure2a} and Fig.\ref{figure2d}, two instances ($\mathbf{x}_{1}$ and $\mathbf{x}_{2}$) not only have a high (low)  feature similarity, but also a high (low) semantic similarity due to largely overlapped (non-overlapped) candidate labels.  From the setting of PML, these two instances are likely to have overlapped (non-overlapped) ground-truth labels. On the other hand, if two instances are without any overlapped candidate label (say zero semantic similarity), their ground-truth labels should be non-overlapped (as Fig. \ref{figure2b} show). Besides, if these two instances have a low feature similarity but with a high semantic similarity (as Fig. \ref{figure2c} show), their ground-truth labels can be overlapped to some extent.

Given these observations, we introduce the Partial Multi-label Learning with Label and Feature Collaboration (PML-LFC). PML-LFC firstly learns a linear predictor with respect to a \emph{latent} ground-truth label matrix, and  induces a low-rank constraint on the coefficient matrix of the predictor to account for the label correlations of multi-label data. Then, it computes the \emph{feature} similarity between instances and the \emph{semantic} similarity between them using the candidate label vectors, and forces the inner product similarity of latent label vectors consistent with the feature similarity and semantic similarity. In this way, both the label and feature information are collaboratively used to induce the latent label vectors, and the four scenarios illustrated in Fig.\ref{figure2} are jointly modelled. PML-LFC finally achieves the predictor and the latent label matrix in a reciprocal reinforce manner by a unified model, and develops an alternative optimization procedure to optimize them.

The main contributions of this paper are summarized as follows:\\
\indent (i) We introduce PML-LFC to jointly leverage the label and feature information of partial multi-label data to induce a credible multi-label classifier, where existing PML solutions isolate the usage of label and feature information, or ignore the usage of negative information that two instances with high (low) feature similarity but with a low (high) semantic similarity from the candidate labels. \\
\indent (ii) PML-LMC unifies the predictor training on PML data and latent label matrix exploration in a unified objective, and introduces an alternative optimization procedure to jointly optimize the predictor and latent label matrix in a mutually beneficial manner.\\
\indent (iii) Empirical study on public multi-label datasets shows that PML-LFC significantly outperforms the related and competitive methods: fPML \cite{yu2018fPML}, PML-LRS \cite{sun2019PML}, DRAMA \cite{wang2019DRAMA}, PRACTICLE \cite{fang2019PRACTICLE}, and two classical multi-label classifiers (RankSVM \cite{elisseeff2002RankSVM}, and ML-KNN \cite{zhang2007mlknn}). In addition, the collaboration between labels and features contributes an improved performance.

The reminder of this paper is organized as follows. Section \ref{sec:relwork} clarifies the difference between our problem and  multi-label learning, partial-label learning, and then reviews the latest PML methods. Section \ref{sec:method} elaborates on the PML-LFC and its optimization procedure. The experimental setup and results are provided and analyzed in Section \ref{sec:exp}. Conclusions and future works are summarized in Section \ref{sec:concl}.

\section{Related Work}
\label{sec:relwork}

PML is different from multi-label crowd consensus learning \cite{li2019mllcrowd,tu2018mljmf,zhang2018kdd}, which wants to obtain high-quality consensus annotations of repeatedly annotated instances, while PML does not have such repeated annotations of the same instances. PML is unlike multi-label weak-label learning \cite{sun2010well,tan2018IJCAI}, which focus on learning from annotated training data with incomplete (missing) labels. PML is also different from the popular partial-label learning (PLL) \cite{cour2011learning,zhang2017disambiguation}, which assumes only one label from the candidate labels of the sample is the ground truth and aims to induce a multi-class predictor to assign one label for unseen sample. PLL can be viewed as a degenerated version of PML.  We observe that PML is more difficult than the typical MLL and PLL problems, since the ground truth labels of samples are not directly accessible to train the predictor and a set of discrete non-exclusive labels should be carefully assigned. To be self-inclusive and help reader being informed, we give a brief review of popular PML solutions.

\cite{xie2018PML} introduced two PML approaches (PML-fp and PML-lc) to elicit the ground-truth labels by minimizing the confidence weighted ranking loss between candidate and non-candidate labels. PML-fp focuses on the utilization of feature information of training data, while PML-lc focuses on the label correlations. To mitigate the negative impact of irrelevant labels in the training phase, \cite{fang2019PRACTICLE} proposed a two-stage approach (PARTICLE), which firstly elicits credible labels via iterative label propagation, and then takes the elicited labels to induce a multi-label classifier with virtual label splitting (PARTICLE-VLS) or maximum a posteriori reasoning (PARTICLE-MAP). \cite{wang2019DRAMA} introduced another two-stage PML approach (DRAMA) that firstly estimates the confidence value for each label by utilizing the feature manifold that indicates how likely a label is correct, and then induces a gradient boosting model to fit the label confidences by exploring the label correlations with the previously elicited labels in each boosting round. Due to the isolation between label elicitation and the classifier training, the elicited labels maybe not compatible with the classifier. \cite{yu2018fPML} introduced a feature-induced PML solution (fPML), which coordinately factorizes the observed sample-label association matrix and the sample-feature matrix into low-rank matrices and then reconstructs the sample-label matrix to identify irrelevant labels. At the same time, fPML optimizes a compatible predictor based on the reconstructed sample-label matrix.  Similarly, \cite{sun2019PML} assumed the observed label matrix is the linear combination of a  ground-truth label matrix with low-rank and an irrelevant label matrix with sparse constraints, and introduced a solution called PML-LRS.

These aforementioned state-of-the-art PML solutions  either mainly focus on the usage of feature manifold that similar instances will have similar labels, or on the usage of ground-truth label matrix being low-rank. They still isolate the joint effect of features and labels for effective partial multi-label learning to some extent.  The (latent) labels of instances are dependent on the features of these instances \cite{zhang2010multilabel,li2012towards}, and the semantic similarity derived from the label sets of multi-label instances are positively correlated with the feature similarity between them \cite{wang2009cvpr,yu2016SimNet,tan2017remotesensing}. Both the label and feature information of multi-label data should be well accounted for effective learning on PML data. Given that, we introduce PML-LFC to collaboratively use the feature and label information, which will be detailed in the next Section.

\section{Proposed method}
\label{sec:method}
\vspace{-1em}
Let $\mathcal{X} \in \mathbb {R}^{d}$ denote the $d$-dimensional feature space, and $\mathcal{L}=\{0|1\}_{c=1}^q$ denote the label space with respect to $q$ distinct labels. Given a PML dataset $\mathcal{D}=\{(\mathbf{x}_i,\mathcal{L}_i)|1 \leq i \leq n \}$, where $\mathbf{x}_i \in \mathcal{X}$ is the feature vector of the $i$-th sample, and $\mathcal{L}_i \subset \mathcal{L}$ is the set of candidate labels currently annotated to $\mathbf{x}_i$. The key characteristic of PML is that only a subset labels $\tilde{\mathcal{L}}_i \subset \mathcal{L}_i$ are the ground-truth labels  of $\mathbf{x}_i$, while the others ($\mathcal{L}_i-\tilde{\mathcal{L}}_i$) are irrelevant for $\mathbf{x}_i$. However, $\tilde{\mathcal{L}}_i$ is not directly accessible to the predictor. The target of PML is to induce a multi-label classifier $F: \mathcal{X}\rightarrow 2^{\mathcal{L}}$ from $\mathcal{D}$.  A naive PML solution is to divide PML problem into $q$ binary sub-problems, and then adopt a noisy label resistant learning algorithm \cite{natarajan2013NIPS,liu2016TPAMI}. But this naive solution generally suffers from the label sparsity issue of multi-label data, where each instance typically is only annotated with several labels of whole label space and each label is only annotated to a small portion of instances. Moreover, it disregards the correlations between labels. Another straightforward solution is to take candidate labels as the ground-truth labels and then apply off-the-shelf MLL algorithms \cite{gibaja2015mllsurvey} to train the predictor. However, the predictor will be seriously misled by the false positive labels in the candidate labels.

To bypass the difficulty of the lack of known ground-truth labels of training instances, we take $\mathbf{P}=\left[\mathbf{p}_{1},\cdots, \mathbf{p}_{n}\right]^{\top}\in[0,1]^{n\times q}$ as the latent label confidence matrix, where $p_{i c}$ reflects the confidence of the $c$-th label being the ground-truth for the $i$-th instance. Unlike existing two-stage approaches \cite{fang2019PRACTICLE,wang2019DRAMA} that firstly estimate the credible labels and then train predictor using the estimated labels. We integrate the estimation of label confidence matrix $\mathbf{P}$ and predictor learning into a unified framework as follows:
\begin{equation}
\begin{aligned}
\begin{split}
\min & \sum_{i=1}^{n}L\left(\mathbf{x}_{i}, \mathbf{p}_{i}, \mathbf{f}\right) + \alpha \Omega\left(\mathbf{f}\right) + \beta \Phi\left(\mathbf{P}\right)
\end{split}
\end{aligned}
\label{Eq1}
\end{equation}
where $L$ denotes the loss function, $\Omega$ controls the complexity of the prediction model $\mathbf{f}$ and $\Phi$ is the regularization term for label confidence matrix $\mathbf{P}$, $\alpha$ and $\beta$ are the trade-off parameters for the last two terms. In this unified formulation, the model is learned from the confidence label matrix $\mathbf{P}$ rather than the original noisy label matrix $\mathbf{Y}$. Therefore, the key is how to obtain reliable confidence matrix $\mathbf{P}$.

In this paper, we propose to train the predictor based on the widely-used the least square loss to fit the confidence label matrix $\mathbf{P}$ as follows:
\begin{equation}
L\left(\mathbf{x}_{i},\mathbf{p}_{i},\mathbf{f}\right) =  \sum_{i=1}^{n}\left\| \mathbf{x}_{i}\mathbf{W} - \mathbf{p}_{i} \right\|^2
\label{Eq2}
\end{equation}
where $\mathbf{W} = \left[\mathbf{w}_{1}, \cdots, \mathbf{w}_{q}\right]^{\top} \in \mathbb{R}^{d \times q}$ is the coefficient matrix for the predictor.
It is recognized the labels of multi-label instances are correlated and the label data matrix of instances should be a low rank one \cite{xu2014learning,yu2018fPML}. Given that,  we instantiate the regularization on the predictor with low-rank constraint on $\mathbf{W}$ as follows:
\begin{equation}
\Omega\left(\mathbf{f}\right) = \texttt{rank}\left(\mathbf{W}\right)
\label{Eq6}
\end{equation}


The main bottleneck of PML problem is the lack of the ground-truth labels of training instances. To overcome this bottleneck, most efforts operate in the feature space based on the assumption that similar (dissimilar) instances have similar (dissimilar) label assignments. They adopt manifold regularization \cite{belkin2006mr} derived from the feature similarity to refine the labels of PML data, and then induce a predictor on the refined labels  \cite{fang2019PRACTICLE,wang2019DRAMA}. Some  efforts work in the label space using the knowledge that the latent ground-truth label matrix should be low-rank \cite{sun2019PML,yu2018fPML}, or that the relevant labels of an instance are hidden in the candidate label set and should be ranked ahead of irrelevant ones outside of the candidate set \cite{xie2018PML}.  Although these efforts leverage the feature information to identify the relevant/irrelevant labels of training instances to some extent, they are still inclined to assign similar labels to two instances with high feature similarity but without any common candidate label. From the definition of PML, it is easy to observe that the ground-truth labels of training instances are hidden in the collected candidate label set. In other words, if two annotated instances do not share any candidate label, then there is no overlap between the individual ground-truth label sets of the two instances (as Fig. \ref{figure2b} show). Besides, they also prefer to assign different label sets to two instances without sufficient large feature similarity but with largely overlapped candidate labels  (as shown in Fig. \ref{figure2c}.
In summary, contemporary PML approaches do not sufficiently use the negative information that  that two instances with high (low) feature similarity but with a low (high) semantic similarity from the candidate labels, since they do not collaboratively use the feature and label information in a coherent way.

To remedy this issue, we specify the last term in Eq. (\ref{Eq1}) as follows:
\begin{equation}
\begin{aligned}
\begin{split}
\Phi\left(\mathbf{P}\right) = &\sum_{i,j, i\neq j} \left( \mathbf{s}_{ij}\mathbf{c}_{ij} - \mathbf{p}_{i}^{\top}\mathbf{p}_{j} \right)^2
\\ &\text{s.t.} \ \mathbf{P} \geq \mathbf{0}, \ \sum_{c=1}^{q} {p}_{ic} = 1,  \forall i = 1,2,3,\cdots,n
\end{split}
\end{aligned}
\label{Eq2}
\end{equation}
where $\mathbf{s}_{ij}$ represents the feature similarity between $\mathbf{x}_{i}$ and $\mathbf{x}_{j}$, $\mathbf{c}_{ij}$ reflects the semantic similarity derived from candidate labels of these two instances, respectively.
The first constraint guarantees that each candidate label has a non-negative confidence value, and the second constraint restricts the confidence value is within [0,1], and the sum of them equal to 1.  We can find that: (i) if two instances have both high values of $\mathbf{s}_{ij}$ and $\mathbf{c}_{ij}$, then $\mathbf{p}_i$ and  $\mathbf{p}_j$ should be close to each other;  (ii) if two instances have a large (or moderate) value of $\mathbf{c}_{ij}$, then $\mathbf{p}_i$ and  $\mathbf{p}_j$ can still have some overlaps; (iii) if two instances have a zero (or low) value of $\mathbf{c}_{ij}$ and a low value of $\mathbf{s}_{ij}$, then $\mathbf{p}_i$ and  $\mathbf{p}_j$ should be not overlapped. Our minimization of Eq. (\ref{Eq2}) jointly considers the above cases. In contrast, contemporary PML methods ignore the semantic similarity between instances. They do not make effective use of the negative information in the last two cases stated above. We want to remark that given the existence of irrelevant labels of training instances, it is not an easy job to quantify and leverage the important label correlation for partial multi-label learning. Thanks to the semantic similarity, which quantifies the similarity between instances based on the pattern that two (or more) labels co-annotate to the same instances, this pattern is also transferred to the latent confident label matrix $\mathbf{P}$. In addition, this pattern transfer is also coordinated by the low-rank constraint on the coefficient matrix and by the feature similarity, which alleviates the negative impact of irrelevant labels on quantifying the semantic similarity. In this way, the information sources from the feature and label spaces are jointly used to guide the latent label matrix learning, which rewards a credible multi-label predictor.

Here, we initialize the label confidence matrix $\mathbf{P}$ as:
\begin{equation}
\forall 1 \leq i \leq n:  \quad p_{i,c} =\left\{\begin{array}{cl}{\frac{1}{\left|\mathcal{L}_i\right|},} & {\text { if } c \in \mathcal{L}_{i}} \\ {0,} & {\text { otherwise }}\end{array}\right.
\label{Eq33}
\end{equation}

To quantify the feature similarity between instances, we adopt the widely-used Gaussian heat kernel similarity as follows:
\begin{equation}
\mathbf{s}_{ij} = exp^{-\frac{\left\|\mathbf{x}_{i} - \mathbf{x}_{j}\right\|_2^2}{t^2}}
\label{Eq3}
\end{equation}
where $t$ denotes the kernel width and is empirically set to $t=\sum_{i,j,i \neq j}^{n} \frac{\left\|\mathbf{x}_{i} - \mathbf{x}_{j}\right\|}{n-1}$. Clearly, $\mathbf{s}_{ij} \in (0,1)$ when there are no two identical instances. We want to remark that other similarity metrics can also be adopted here. Our choice of Gaussian heat kernel is for its simplicity and wide application.

Diverse similarity metrics can also be adopted to quantify the semantic similarity between multi-label instances \cite{wang2009cvpr,yu2016SimNet}, here we use the cosine similarity as follows:
\begin{equation}
\mathbf{c}_{ij} = \frac{\mathbf{y}_{i}^{\top} \mathbf{y}_{j} }{\left\|\mathbf{y}_{i}\right\|\left\|\mathbf{y}_{j}\right\|}
\label{Eq4}
\end{equation}
where $\mathbf{y}_i$ is the one-hot coding label vector for $\mathbf{x}_i$, $\mathbf{y}_{ic}=1$ if the $c$-th label is annotated to $\mathbf{x}_i$,  $\mathbf{y}_{ic}=0$ otherwise. Obviously, $\mathbf{c}_{ij}\in [0,1]$, it has a large value when two instances have a large portion of overlapped candidate labels, moderate value when they share some overlapped candidate labels, and zero value when they do not have any overlapped candidate label.


Based on the above analysis, we can instantiate the PML-LFC as follows:
\begin{equation}
\begin{aligned}
\begin{split}
\min_{\mathbf{W}, \mathbf{P} \geq \mathbf{0}} & \left\| \mathbf{XW} - \mathbf{P} \right\|_{2}^2 + \alpha \texttt{rank}(\mathbf{W}) + \beta\sum_{i,j, i\neq j} \left( \mathbf{s}_{ij}\mathbf{c}_{ij} - \mathbf{p}_{i}^{\top}\mathbf{p}_{j} \right)^2
\\& \text { s.t. } \sum_{c=1}^{q} {p}_{ic} = 1,  \forall i = 1,2,3,\cdots,n
\end{split}
\end{aligned}
\end{equation}
  The problem above can be further rewritten as follows:
\begin{equation}
\begin{aligned}
\begin{split}
\min_{\mathbf{W}, \mathbf{P} \geq \mathbf{0}} & \left\| \mathbf{XW} - \mathbf{P} \right\|_{2}^2 + \alpha \texttt{rank}(\mathbf{W}) + \beta \left\| \mathbf{H} \odot \left(\mathbf{S} \odot \mathbf{C} - \mathbf{P}\mathbf{P}^\top\right) \right\|_{2}^2
\\& \text { s.t. } \sum_{c=1}^{q} {p}_{ic} = 1,  \forall i = 1,2,3,\cdots,n
\end{split}
\end{aligned}
\label{Eq77}
\end{equation}
where $\mathbf{H} \in \mathbb{R}^{n\times n}$, $\mathbf{H}_{ij} = 0$ if $i = j$; and $\mathbf{H}_{ij} = 1$ otherwise, $\odot$ is the Hadamard product. However, the rank function in Eq.(\ref{Eq77}) is hard to optimize, the nuclear norm $\left\| \bullet \right\|_{*}$ is suggested to surrogate the rank function. Therefore, Eq.(\ref{Eq77}) is reformulated as follows:
\begin{equation}
\begin{aligned}
\begin{split}
\min_{\mathbf{W}, \mathbf{P} \geq \mathbf{0}} & \left\| \mathbf{XW} - \mathbf{P} \right\|_{2}^2 + \alpha \|\mathbf{W}\|_{*}+ \beta \left\| \mathbf{H} \odot \left(\mathbf{S} \odot \mathbf{C} - \mathbf{P}\mathbf{P}^\top\right) \right\|_{2}^2
\\& \text { s.t. } \sum_{c=1}^{q} {p}_{ic} = 1,  \forall i = 1,2,3,\cdots,n
\end{split}
\end{aligned}
\label{Eq7}
\end{equation}




\subsection{Optimization}
\vspace{-1em}
Since the optimization problem in Eq.(\ref{Eq7}) is non-convex with respect to $\mathbf{W}$ and $\mathbf{P}$ at the same time. We apply the alternative optimization procedure to approximate them. Specifically, we alternatively optimize one variable while fixing the other one as a constant. The detailed procedure is presented below.

\textbf{Update W}: With $\mathbf{P}$ fixed, Eq.(\ref{Eq7})  with respect to $\mathbf{W}$ is equivalent to the following problem:
\begin{equation}
\min_{\mathbf{W}} \left\| \mathbf{XW} - \mathbf{P} \right\|_{2}^2 + \alpha \left\| \mathbf{W} \right\|_{*}
\label{Eq8}
\end{equation}
The minimization of Eq. (\ref{Eq8}) is a trace norm minimization problem, which is time-consuming. To reduce the computation time of Eq. (\ref{Eq8}), we use the Accelerated Gradient Descent (AGD)  algorithm \cite{ji2009accelerated} to optimize $\mathbf{W}$ as summarized in Algorithm \ref{Algorithm1}.

In particular, $F\left(\mathbf{W}\right)$, $p_{l_{t}}\left(\mathbf{Z}_{t}\right)$ and $Q_{l_{t}}\left(\mathbf{W}_{t},\mathbf{Z}_{t}\right)$ in Algorithm \ref{Algorithm1} are defined as follows:
\begin{equation}
    F\left(\mathbf{W}\right) = \left\| \mathbf{XW} - \mathbf{P} \right\|_{2}^2 + \alpha \left\| \mathbf{W} \right\|_{*}
\end{equation}
\begin{equation}
    p_{l_{t}}\left(\mathbf{Z}_{t}\right) = \frac{l_{t}}{2}\left\|\mathbf{W} - \left(\mathbf{Z}_{t} - \frac{1}{l_{t}}\nabla f\left(\mathbf{Z}_{t}\right)\right)\right\|_{2}^2 + \alpha\left\|\mathbf{W}\right\|_{*}
\end{equation}
\begin{equation}
    f\left(\mathbf{Z}_{t}\right) = \left\|\mathbf{Z}_{t}\mathbf{X}-\mathbf{P}\right\|_{2}^2
\end{equation}
\begin{equation}
    Q_{l_{t}}\left(\mathbf{W}_{t},\mathbf{Z}_{t}\right)=f\left(\mathbf{Z}_{t}\right) + \left<\mathbf{W}_{t} - \mathbf{Z}_{t}, \nabla f\left(\mathbf{Z}_{t}\right)\right>+\frac{l_{t}}{2}\left\|\mathbf{W}_{t}-\mathbf{Z}_{t}\right\|_{2}^2 + \alpha \left\|\mathbf{W}\right\|_{*}
\end{equation}

\textbf{Update P}: With $\mathbf{W}$ fixed, Eq.(\ref{Eq7}) with respect to $\mathbf{P}$ reduces to:
\begin{equation}
\begin{aligned}
\begin{split}
\min_{\mathbf{P} \geq \mathbf{0}} & \left\| \mathbf{XW} - \mathbf{P} \right\|_{2}^2 + \beta \left\| \mathbf{H} \odot \left(\mathbf{S} \odot \mathbf{C} - \mathbf{P}\mathbf{P}^\top\right) \right\|_{2}^2
\\& \text { s.t. } \sum_{c=1}^{q} {p}_{ic} = 1,  \forall i = 1,2,3,...,n
\end{split}
\end{aligned}
\label{Eq10}
\end{equation}
By introducing the Lagrange multiplier $\lambda$, Eq. (\ref{Eq10}) is equivalent to:
\begin{equation}
\min_{\mathbf{P} \geq \mathbf{0}}  \left\| \mathbf{XW} - \mathbf{P} \right\|_{2}^2 + \beta \left\| \mathbf{H} \odot \left(\mathbf{A} - \mathbf{P}\mathbf{P}^\top\right) \right\|_{2}^2 + \lambda \left\|\mathbf{P}\mathbf{1}_{q} - \mathbf{1}_{n}\right\|_{2}^2
\label{Eq14}
\end{equation}
where $\mathbf{A}=\mathbf{S} \odot \mathbf{C}$, $\mathbf{1}_{q}$ ($\mathbf{1}_{n}$) are the $q$($n$)-dimensional column vector with all ones.
The gradient with respect to $\mathbf{P}$ is:
\begin{equation}
\begin{aligned}
\begin{split}
           \nabla \mathbf{P} =& \mathbf{P}-\mathbf{XW} - \beta \mathbf{H} \odot \left(\mathbf{A} - \mathbf{P}\mathbf{P}^\top\right)\mathbf{P} \\
        & -  \beta \mathbf{H}^\top \odot \left(\mathbf{A}^\top
        - \mathbf{P}\mathbf{P}^\top \right)\mathbf{P} + \lambda\left(\mathbf{P}\mathbf{1}_{q}-\mathbf{1}_{n}\right)\mathbf{1}_{q}^\top
\end{split}
\end{aligned}
\end{equation}
We can use the Karush-Kuhn-Tucker (KKT) conditions \cite{boyd2004convex} for the nonnegativity of $\mathbf{P}$ as:
\begin{equation}
\begin{aligned}
\begin{split}
    &(\mathbf{P} - \mathbf{XW} - \beta \mathbf{H} \odot \mathbf{AP} + \beta \mathbf{H}\odot\mathbf{B} - \beta\mathbf{H}^\top \odot \mathbf{A}^\top \mathbf{P}
    \\&+ \beta \mathbf{H}^\top\odot\mathbf{B} + \lambda \mathbf{P}\mathbf{Q} - \lambda \mathbf{N})_{ij} \mathbf{P}_{ij}=0
\end{split}
\end{aligned}
\label{Eq16}
\end{equation}
where $\mathbf{B} = \mathbf{P}\mathbf{P}^\top\mathbf{P}$, $\mathbf{Q}$ and $\mathbf{N}$ are all-one matrices with $q \times q$ dimensions and $n \times q$ dimensions, respectively. Let $\mathbf{XW} = \mathbf{XW}^+ - \mathbf{XW}^-$, where $\mathbf{M}_{ij}^{+} = \frac{|\mathbf{M}|_{ij} + \mathbf{M}_{ij}}{2}$ and $\mathbf{M}_{ij}^{-} = \frac{|\mathbf{M}|_{ij} - \mathbf{M}_{ij}}{2}$ for any matrix $\mathbf{M}$, Eq. (\ref{Eq16}) can be rewritten as:
\begin{equation}
\begin{aligned}
\begin{split}
&(\mathbf{P} - \mathbf{XW}^+ + \mathbf{XW}^- - \beta \mathbf{H} \odot \mathbf{AP} + \beta \mathbf{H}\odot\mathbf{B} - \beta\mathbf{H}^\top \odot \mathbf{A}^\top \mathbf{P}
    \\&+ \beta \mathbf{H}^\top\odot\mathbf{B} + \lambda \mathbf{P}\mathbf{Q} - \lambda \mathbf{N})_{ij} \mathbf{P}_{ij}=0
\end{split}
\end{aligned}
\label{Eq17}
\end{equation}
Eq. ({\ref{Eq17}}) leads to the following update formula:
\begin{equation}
\begin{aligned}
\begin{split}
\mathbf{P}_{ij} = \mathbf{P}_{ij}\sqrt{\frac{(\mathbf{XW}^{+} + \beta \mathbf{H} \odot \mathbf{AP} + \beta \mathbf{H}^\top \odot \mathbf{A}^\top\mathbf{P} + \lambda\mathbf{N})_{ij}}{(\mathbf{P} + \mathbf{XW}^{-} + \beta \mathbf{H} \odot \mathbf{B} + \beta \mathbf{H}^\top \odot \mathbf{B} + \lambda \mathbf{PQ})_{ij}}}
\end{split}
\end{aligned}
\end{equation}

Algorithm \ref{Algorithm2} summarizes the pseudo-code of PML-LFC. We observe that PML-LFC only needs at most ten iterations to converge on our used datasets.

\begin{algorithm}[h!tbp]
\caption{Optimization of $\mathbf{W}$}
\hspace*{0.02in} {\bf Input:}
$\mathbf{X}$, $\mathbf{P}$, $\alpha$.\\
\hspace*{0.02in} {\bf Output:}
$\mathbf{W}$.
\begin{algorithmic}[1]
\State Initialize $\mathbf{W}_{0}$ = $\mathbf{Z} \in \mathbb{R}^{d \times q}$, $l_{0} > 0$,$\gamma >0$, $\delta_{1} = 1$, $t=1$
\State{\bf Iterate:}
\State\quad Set $\bar{l} = l_{t-1}$
\State\quad {\bf While} $F\left(p_{\bar{l}}\left(\mathbf{Z}_{t-1}\right)\right) > Q_{\bar{l}}\left(p_{\bar{l}}\left(\mathbf{Z}_{t-1}\right),\mathbf{Z}_{t-1}\right)$
\State\qquad Set $\bar{l} = \gamma\bar{l}$
\State\qquad Set $l_{t} = \bar{l}$ and update
\State\qquad $\mathbf{W}_{t} = p_{l_{t}}\left(\mathbf{Z}_{t}\right)$
\State\qquad $\delta_{t+1} = \frac{ 1 + \sqrt{1 + 4\delta_{t}^2}}{2}$
\State\qquad $\mathbf{Z}_{t+1} = \mathbf{W}_{t} + \left(\frac{\delta_{t} -1}{\delta_{t} + 1}\right)\left(\mathbf{W}_{t} - \mathbf{W}_{t+1}\right)$
\end{algorithmic}
\label{Algorithm1}
\end{algorithm}

\vspace{-1em}
\begin{algorithm}[t]
\caption{PML-LFC: Partial Multi-label Learning with Label and Feature Collaboration}
\hspace*{0.02in} {\bf Input:} \\
\hspace*{0.35in} {$\mathbf{X}$: $n \times d$ instance-feature matrix;\\}
\hspace*{0.1in} {\qquad $\mathbf{Y}$: $n \times q$ instance-label association matrix;\\}
\hspace*{0.1in} {\qquad $\alpha$, $\beta$: scalar input parameters.\\}
\hspace*{0.02in} {\bf Output:} \\
\hspace*{0.1in}  {\qquad Prediction coefficients $\mathbf{W}$.}
\begin{algorithmic}[1]
\State Initialize $\mathbf{P}$ by Eq.(\ref{Eq33});
\State{\bf Do:}
\State \quad Seek the optimal $\mathbf{W}$ by optimizing Eq. (\ref{Eq8}) and Algorithm \ref{Algorithm1};
\State \quad Fix $\mathbf{W}$, update $\mathbf{P}$ by optimizing Eq. (\ref{Eq10});
\State{\bf While} not convergence or within the allowed number of iterations
\end{algorithmic}
\label{Algorithm2}
\end{algorithm}
\vspace{-1em}
\section{Experiments}
\label{sec:exp}
\vspace{-1em}
\subsection{Experimental Setup}
\vspace{-1em}
\textbf{Dataset:} For a quantitative performance evaluation, five synthetic and three real-world PML datasets are collected for experiments. Table \ref{table1} summarizes characteristics of these datasets. Specifically, to create a synthetic PML dataset, we take the current labels of instances as ground-truth ones. For each instance $\mathbf{x}_{i}$, we randomly insert the irrelevant labels of $\mathbf{x}_{i}$ with ${a}\%$ number of the ground-truth labels, and we vary ${a}\%$ in the range of $\{ 10 \%, 50 \%, 100 \%, 200 \% \}$. All the datasets are randomly partitioned into 80\% for training and the rest 20\% for testing. We repeat all the experiments for 10 times independently, report the average results with standard deviations.\\
\textbf{Comparing methods:} Four representative PML algorithms, including fPML \cite{yu2018fPML}, PML-LRS \cite{sun2019PML}, DRAMA \cite{wang2019DRAMA} and PARTICLE-VLS \cite{fang2019PRACTICLE} are used as the comparing methods. DRAMA and PARTICLE-VLS mainly utilize the feature similarity between instances, while fPML and PML-LRS build on low-rank assumption of the label matrix, and fPML additionally explores and uses the coherence between the label and feature data matrix. In addition, two representative MLL solutions (ML-KNN \cite{zhang2007mlknn} and Rank-SVM\cite{elisseeff2002RankSVM}) are also included as baselines for comparative analysis. The last two comparing methods directly take the candidate labels as ground-truths to train the respective predictor. For these comparing methods, parameter configurations are fixed or optimized by the suggestions in the original codes or papers. For our PML-LMC, we fix $\alpha$=10 and $\beta$=10. The parameter sensitivity of $\alpha$ and $\beta$ will be analyzed later.\\
\textbf{Evaluation metrics:} For a comprehensive performance evaluation and comparison, we adopt five widely-used multi-label evaluation metrics: \textit{hamming loss} (HammLoss), \textit{one-error} (OneError), \textit{coverage} (Coverage), \textit{ranking loss} (RankLoss) and \textit{average precision}(AvgPrec). The formal definition of these metrics can be founded in \cite{zhang2014mllreview,gibaja2015mllsurvey}. Note here \textit{coverage} is normalized by the number of distinct labels, thus it ranges in [0,1]. Furthermore, the larger the value of \textit{average precision}, the better the performance is, while the opposite holds for the other four evaluation metrics.
\vspace{-1em}
\begin{table*}[h!tb]
\renewcommand\arraystretch{0.7} 
\scriptsize
\vspace{-1em}
\centering
\caption{Characteristics of the experimental dataset. `\#Instance' is the number of Examples, `\#Features' is the number of features, `\#Labels' is the number of distinct labels, `avgGLs' is the average number of ground-truth labels of each instance, and `\#' is the number of noise labels of the dataset.}
\begin{tabular}{c r r r r r}
    \hline
    \hline
        Data set\qquad & $\#$Instances \qquad & $\#$Features \qquad& $\#$Labels \qquad& avgGLs \qquad& $\#$Noise\\
    \midrule
        slashdot & 3782 & 1079 & 22 & 1.181 & -\\
        scene & 2407 & 294 & 6 & 1.074 & -\\
        enron & 1702 & 1001 & 53 & 3.378 & -\\
        medical & 978 & 1449 & 45 & 1.245 & -\\
        Corel5k & 5000 & 499 & 374 & 0.245 & - \\
\hline
\hline
        \emph{YeastBP} & 6139 & 6139 & 217 & 5.537 &2385\\
        \emph{YeastCC} & 6139& 6139 &50 & 1.348 &260\\
         \emph{YeastMF}& 6139 & 6139 & 39 & 1.005 &234\\
\hline
\hline
\end{tabular}
\label{table1}
\end{table*}
\subsection{Results and Analysis}
Table \ref{table2} reports the detailed experimental results of six comparing algorithms with the noisy label ratio of 50\%, while similar observation can be found in terms of other noisy label ratios. 
The first stage of DARAM and PARTLCE-VLS utilizes the feature similarity to optimize the ground-truth label confidence in different ways. However, due to the features of three real-world PML datasets are protein-protein interaction networks, we directly use the network structure to optimize the ground-truth label confidence matrix in the first stage by respective algorithms. In the second stage, PARTICLE-VLS introduces a virtual label technique to transform the problem into multiple binary training sets, and results in the class-imbalanced problem and causes computation exception due to the sparse biological network data. For this reason, its results on the last three datasets can not be reported. Due to page limit, we  summarize the win/tie/loss
counts of our method versus the other comparing method in 23 cases (five datasets $\times$ four ratios of noisy labels  and three PML datasets) across five evaluation metrics in  Table \ref{table3}.

\begin{table*}[h!tbp]
\renewcommand\arraystretch{0.7} 
\centering
\scriptsize
    \caption{Experiment results on different datasets with noisy labels (50\%). $\bullet$/$\circ$ indicates whether PML-LFC is statistically (according to pairwise $t$-test at 95\% significance level) superior/inferior to the other method.}
     \begin{center}
    \begin{tabular}{c|c c| c c c c| c }
    \hline
    {Metric} &RankSVM &ML-KNN &PML-LRS &fPML &DARAM &PARTICLE-VLS &PML-LFC \\
    \hline
     & \multicolumn{7}{c}{slashdot}\\
     \cline{2-8}
     HammLoss &0.078$\pm$0.005$\bullet$
     &0.184$\pm$0.006$\bullet$
     &0.044$\pm$0.000$\circ$
     &0.043$\pm$0.000$\circ$
     &0.052$\pm$0.000$\circ$
     &0.053$\pm$0.001$\circ$
     &0.073$\pm$0.000\\
     RankLoss &0.161$\pm$0.002$\bullet$
     &0.053$\pm$0.000$\circ$
     &0.153$\pm$0.006$\bullet$
     &0.127$\pm$0.006$\bullet$
     &0.118$\pm$0.000$\bullet$
     &0.305$\pm$0.032$\bullet$
     &0.110$\pm$0.007\\
     OneError &0.534$\pm$0.005$\bullet$
     &0.680$\pm$0.015$\bullet$
     &0.446$\pm$0.012$\bullet$
     &0.480$\pm$0.013$\bullet$
     &0.404$\pm$0.001$\bullet$
     &0.769$\pm$0.074$\bullet$
     &0.393$\pm$0.028\\
     Coverage&0.182$\pm$0.022$\bullet$
     &0.120$\pm$0.006$\circ$
     &0.165$\pm$0.007$\bullet$
     &0.139$\pm$0.003$\bullet$
     &0.133$\pm$0.001$\bullet$
     &0.305$\pm$0.031$\bullet$
     &0.128$\pm$0.007\\
     AvgPrec&0.582$\pm$0.017$\bullet$
     &0.472$\pm$0.011$\bullet$
     &0.639$\pm$0.007$\bullet$
     &0.627$\pm$0.007$\bullet$
     &0.686$\pm$0.001$\bullet$
     &0.375$\pm$0.036$\bullet$
     &0.696$\pm$0.010\\
     \hline
     & \multicolumn{7}{c}{scene}\\
    \cline{2-8}
     HammLoss &0.272$\pm$0.012$\bullet$
     &0.110$\pm$0.013$\circ$
     &0.148$\pm$0.005$\bullet$
     &0.167$\pm$0.001$\bullet$
     &0.121$\pm$0.000$\circ$
     &0.123$\pm$0.017$\circ$
     &0.146$\pm$0.003\\
     RankLoss &0.259$\pm$0.015$\bullet$
     &0.097$\pm$0.008$\bullet$
     &0.124$\pm$0.011$\bullet$
     &0.145$\pm$0.005$\bullet$
     &0.118$\pm$0.002$\bullet$
     &0.110$\pm$0.020$\bullet$
     &0.094$\pm$0.004\\
     OneError &0.553$\pm$0.009$\bullet$
     &0.260$\pm$0.009$\bullet$
     &0.314$\pm$0.027$\bullet$
     &0.362$\pm$0.009$\bullet$
     &0.265$\pm$0.003$\bullet$
     &0.251$\pm$0.044$\circ$
     &0.258$\pm$0.007\\
     Coverage &0.232$\pm$0.017$\bullet$
     &0.109$\pm$0.011$\bullet$
     &0.118$\pm$0.009$\bullet$
     &0.136$\pm$0.006$\bullet$
     &0.114$\pm$0.001$\bullet$
     &0.097$\pm$0.018$\bullet$
     &0.093$\pm$0.002\\
    AvgPrec &0.635$\pm$0.038$\bullet$
     &0.838$\pm$0.016$\bullet$
     &0.804$\pm$0.016$\bullet$
     &0.774$\pm$0.005$\bullet$
     &0.830$\pm$0.001$\bullet$
     &0.828$\pm$0.033$\bullet$
     &0.843$\pm$0.005\qquad\\
     \hline
      &\multicolumn{7}{c}{enorn}\\
    \cline{2-8}
     HammLoss &0.109$\pm$0.006$\bullet$
     &0.108$\pm$0.006$\bullet$
     &0.060$\pm$0.001$\bullet$
     &0.104$\pm$0.002$\bullet$
     &0.068$\pm$0.001$\bullet$
     &0.064$\pm$0.008$\bullet$
     &0.051$\pm$0.001\\
     RankLoss &0.189$\pm$0.037$\bullet$
     &0.054$\pm$0.000$\circ$
     &0.145$\pm$0.009$\bullet$
     &0.197$\pm$0.009$\bullet$
     &0.143$\pm$0.002$\bullet$
     &0.238$\pm$0.037$\bullet$
     &0.099$\pm$0.008\\
     OneError &0.476$\pm$0.047$\bullet$
     &0.323$\pm$0.032$\bullet$
     &0.326$\pm$0.036$\bullet$
     &0.416$\pm$0.030$\bullet$
     &0.260$\pm$0.004$\bullet$
     &0.453$\pm$0.102$\bullet$
     &0.254$\pm$0.013\\
     Coverage &0.481$\pm$0.038$\bullet$
     &0.285$\pm$0.005$\bullet$
     &0.369$\pm$0.014$\bullet$
     &0.331$\pm$0.016$\bullet$
     &0.354$\pm$0.002$\bullet$
     &0.451$\pm$0.071$\bullet$
     &0.284$\pm$0.010\\
     AvgPrec &0.504$\pm$0.053$\bullet$
     &0.611$\pm$0.019$\bullet$
     &0.613$\pm$0.015$\bullet$
     &0.659$\pm$0.008$\bullet$
     &0.613$\pm$0.002$\bullet$
     &0.466$\pm$0.088$\bullet$
     &0.683$\pm$0.009\\
     \hline
     & \multicolumn{7}{c}{medical}\\
        \cline{2-8}
     HammLoss &0.482$\pm$0.008$\bullet$
     &0.070$\pm$0.009$\bullet$
     &0.343$\pm$0.034$\bullet$
     &0.022$\pm$0.002$\circ$
     &0.015$\pm$0.000$\circ$
     &0.021$\pm$0.001$\circ$
     &0.024$\pm$0.000\\
     RankLoss &0.018$\pm$0.003$\circ$
     &0.042$\pm$0.006$\bullet$
     &0.075$\pm$0.027$\bullet$
     &0.046$\pm$0.005$\bullet$
     &0.036$\pm$0.003
     &0.113$\pm$0.021$\bullet$
     &0.036$\pm$0.005\\
     OneError &0.169$\pm$0.004$\circ$
     &0.270$\pm$0.020$\bullet$
     &0.420$\pm$0.013$\bullet$
     &0.216$\pm$0.008$\bullet$
     &0.193$\pm$0.008$\circ$
     &0.220$\pm$0.082$\bullet$
     &0.199$\pm$0.013\\
     Coverage &0.276$\pm$0.025$\bullet$
     &0.095$\pm$0.011$\bullet$
     &0.114$\pm$0.027$\bullet$
     &0.065$\pm$0.010$\bullet$
     &0.058$\pm$0.001$\bullet$
     &0.116$\pm$0.020$\bullet$
     &0.052$\pm$0.009\\
     AvgPrec &0.854$\pm$0.024$\circ$
     &0.766$\pm$0.015$\bullet$
     &0.665$\pm$0.018$\bullet$
     &0.831$\pm$0.007$\bullet$
     &0.839$\pm$0.007$\circ$
     &0.730$\pm$0.022$\bullet$
     &0.834$\pm$0.012\\
     \hline
     & \multicolumn{7}{c}{Corel5k}\\
             \cline{2-8}
     HammLoss &0.081$\pm$0.007$\bullet$
     &0.161$\pm$0.005$\bullet$
     &0.051$\pm$0.009$\bullet$
     &0.010$\pm$0.000
     &0.554$\pm$0.000$\bullet$
     &0.019$\pm$0.000$\bullet$
     &0.010$\pm$0.000\\
     RankLoss &0.281$\pm$0.006$\bullet$
     &0.134$\pm$0.000$\bullet$
     &0.063$\pm$0.005$\circ$
     &0.210$\pm$0.008$\bullet$
     &0.277$\pm$0.001$\bullet$
     &0.326$\pm$0.056$\bullet$
     &0.120$\pm$0.006\\
     OneError &0.802$\pm$0.007$\bullet$
     &0.740$\pm$0.010$\bullet$
     &0.639$\pm$0.017$\bullet$
     &0.649$\pm$0.008$\bullet$
     &0.801$\pm$0.002$\bullet$
     &0.855$\pm$0.073$\bullet$
     &0.631$\pm$0.010\\
     Coverage &0.391$\pm$0.007$\bullet$
     &0.372$\pm$0.010$\bullet$
     &0.403$\pm$0.007$\bullet$
     &0.470$\pm$0.017$\bullet$
     &0.539$\pm$0.003$\bullet$
     &0.547$\pm$0.041$\bullet$
     &0.281$\pm$0.013\\
     AvgPrec &0.292$\pm$0.008$\bullet$
     &0.230$\pm$0.003$\bullet$
     &0.393$\pm$0.006$\circ$
     &0.286$\pm$0.005$\bullet$
     &0.199$\pm$0.008$\bullet$
     &0.144$\pm$0.052$\bullet$
     &0.312$\pm$0.002\\
     \hline
     & \multicolumn{7}{c}{YeastBP}\\
             \cline{2-8}
    HammLoss&--
    &0.316$\pm$0.005$\bullet$
    &0.329$\pm$0.012$\bullet$
    &0.071$\pm$0.004$\bullet$
    &0.062$\pm$0.000$\bullet$
    &--
    &0.024$\pm$0.000\\
    RankLoss&--
    &0.025$\pm$0.000$\circ$
    &0.331$\pm$0.007$\bullet$
    &0.208$\pm$0.009$\bullet$
    &0.161$\pm$0.000$\bullet$
    &--
    &0.143$\pm$0.002\\
    OneError&--
    &0.757$\pm$0.008$\bullet$
    &0.743$\pm$0.013$\bullet$
    &0.682$\pm$0.004$\bullet$
    &0.796$\pm$0.002$\bullet$
    &--
    &0.523$\pm$0.013\\
    Coverage&--
    &0.407$\pm$0.010$\bullet$
    &0.374$\pm$0.008$\bullet$
    &0.312$\pm$0.005$\bullet$
    &0.295$\pm$0.002$\bullet$
    &--
    &0.281$\pm$0.012\\
    AvgPrec&--
    &0.232$\pm$0.007$\bullet$
    &0.242$\pm$0.011$\bullet$
    &0.394$\pm$0.012$\bullet$
    &0.214$\pm$0.001$\bullet$
    &--
    &0.411$\pm$0.012\\
    \hline
     & \multicolumn{7}{c}{YeastCC}\\
             \cline{2-8}
    HammLoss&0.046$\pm$0.008$\bullet$
    &0.318$\pm$0.016$\bullet$
    &0.351$\pm$0.012$\bullet$
    &0.093$\pm$0.005$\bullet$
    &0.071$\pm$0.000$\bullet$
    &--
    &0.027$\pm$0.000\\
    RankLoss&0.188$\pm$0.004$\bullet$
    &0.026$\pm$0.000$\circ$
    &0.308$\pm$0.009$\bullet$
    &0.179$\pm$0.007$\bullet$
    &0.178$\pm$0.000$\bullet$
    &--
    &0.173$\pm$0.008\\
    OneError&0.555$\pm$0.004$\bullet$
    &0.639$\pm$0.018$\bullet$
    &0.658$\pm$0.014$\bullet$
    &0.524$\pm$0.007$\bullet$
    &0.832$\pm$0.003$\bullet$
    &--
    &0.448$\pm$0.014\\
    Coverage&0.107$\pm$0.009$\bullet$
    &0.173$\pm$0.010$\bullet$
    &0.150$\pm$0.007$\bullet$
    &0.112$\pm$0.004$\bullet$
    &0.111$\pm$0.002$\bullet$
    &--
    &0.103$\pm$0.003\\
    AvgPrec&0.516$\pm$0.010$\bullet$
    &0.398$\pm$0.018$\bullet$
    &0.386$\pm$0.012$\bullet$
    &0.535$\pm$0.009$\bullet$
    &0.193$\pm$0.002$\bullet$
    &--
    &0.590$\pm$0.014\\
    \hline
     & \multicolumn{7}{c}{YeastMF}\\
             \cline{2-8}
    HammLoss&0.055$\pm$0.005$\bullet$
    &0.338$\pm$0.004$\bullet$
    &0.348$\pm$0.004$\bullet$
    &0.044$\pm$.008$\bullet$
    &0.077$\pm$0.001$\bullet$
    &--
    &0.026$\pm$0.000\\
    RankLoss&0.253$\pm$0.009$\bullet$
    &0.025$\pm$0.000$\circ$
    &0.386$\pm$0.008$\bullet$
    &0.269$\pm$0.006$\bullet$
    &0.251$\pm$0.000$\bullet$
    &--
    &0.243$\pm$0.012\\
    OneError&0.681$\pm$0.010$\bullet$
    &0.785$\pm$0.005$\bullet$
    &0.761$\pm$0.012$\bullet$
    &0.693$\pm$0.009$\bullet$
    &0.878$\pm$0.001$\bullet$
    &--
    &0.661$\pm$0.017\\
    Coverage&0.123$\pm$0.008$\bullet$
    &0.172$\pm$0.006$\bullet$
    &0.168$\pm$0.007$\bullet$
    &0.124$\pm$0.003$\bullet$
    &0.137$\pm$0.001$\bullet$
    &--
    &0.121$\pm$0.005\\
    AvgPrec&0.421$\pm$0.008$\bullet$
    &0.330$\pm$0.006$\bullet$
    &0.302$\pm$0.010$\bullet$
    &0.442$\pm$0.009$\bullet$
    &0.160$\pm$0.000$\bullet$
    &--
    &0.457$\pm$0.011\\
    \hline
    \end{tabular}
    \end{center}
\label{table2}
\end{table*}

\begin{table*}[h!tb]
\renewcommand\arraystretch{0.7} 
\centering
\scriptsize
\caption{Win/Tie/Lose counts (pairwise $t$-test at 95\% signification level) of PML-LFC against each other comparing algorithm with different ratios of noisy labels \{10\%, 50\%, 100\%, 200\%\} on different datasets across five evaluation criteria.}
     \begin{center}
     \begin{tabular}{c|c c|c c c c}
     \hline
     \hline
     \multirow{2}*{Metric} & \multicolumn{6}{c}{PML-LFC against}\\
     \cline{2-7}
    &RankSVM &ML-KNN &PML-LRS &fPML & DARAM & PARTICLE-VLS\\
    \hline
     HammLoss & 21/0/2
     &17/2/4
     &18/2/3
     &16/3/4
     &16/2/5
     &15/3/5
     \\RankLoss &20/1/2
     &16/2/5
     &22/1/0
     &19/1/3
     &18/2/3
     &22/0/1
     \\OneError&22/0/1
     &23/0/0
     &23/0/0
     &21/0/2
     &19/0/4
     &20/0/3
     \\Coverage&21/1/1
     &23/0/0
     &22/0/1
     &23/0/0
     &21/1/1
     &20/0/3
     \\AvgPrec&22/0/1
     &23/0/0
     &20/1/2
     &19/1/3
     &19/0/4
     &21/0/2\\
     \hline
     \hline
     Total (Win/Tie/Lose)&106/2/7
     &102/4/9
     &105/4/6
     &98/5/12
     &93/5/17
     &98/3/14\\
     \hline
     \hline
     \end{tabular}
     \end{center}
\label{table3}
\end{table*}

Based on the results in Table \ref{table2} and \ref{table3}, we can observe the following:
(i) On the real-word PML datasets \emph{YeastBP},\emph{YeastCC} and \emph{YeastMF}, PML-LFC achieve the best performance in most cases except ML-KNN on \emph{ranking loss} evaluation. (ii) On the synthic datasets, PML-LFC frequently outperforms other methods and slightly loses to RankSVM and DARAM on medical dataset. (iii) Out of 115 statistical tests
PML-LFC achieves much better results than the popular PML methods PML-LRS, fPML, DARAM and PARTICLE-VLS in 91.30\%, 85.22\%, 80.87\% and 85.22\% cases, respectively.
PML-LFC also significantly outperforms two classical MLL approaches RankSVM and ML-KNN in 92.17\% and 88.70\% cases, respectively. Which proves the necessity of accounting for irrelevant labels of PML training data.
PML-LFC outperforms PML-LRS in most cases because PML-LRS mainly operates in the label space. fPML is similar to PML-LRS, while it uses feature information to guide the low-rank label matrix approximation. As a result, it sometimes obtains similar results as PML-LFC.  PML-LFC also performs better than DARAM and PARTICLE-VLS, which mainly use information from the feature space. Another cause for the superiority of PML-LFC is that other comparing methods do not make a concrete use of the negative information between the label and feature space. From these results, we can draw a conclusion that PML-LFC well accounts the negative information between features and labels for effective partial multi-label learning.

\subsection{Further Analysis}
\begin{figure}[h!t]
\centering
\includegraphics[width=9cm,height=5cm]{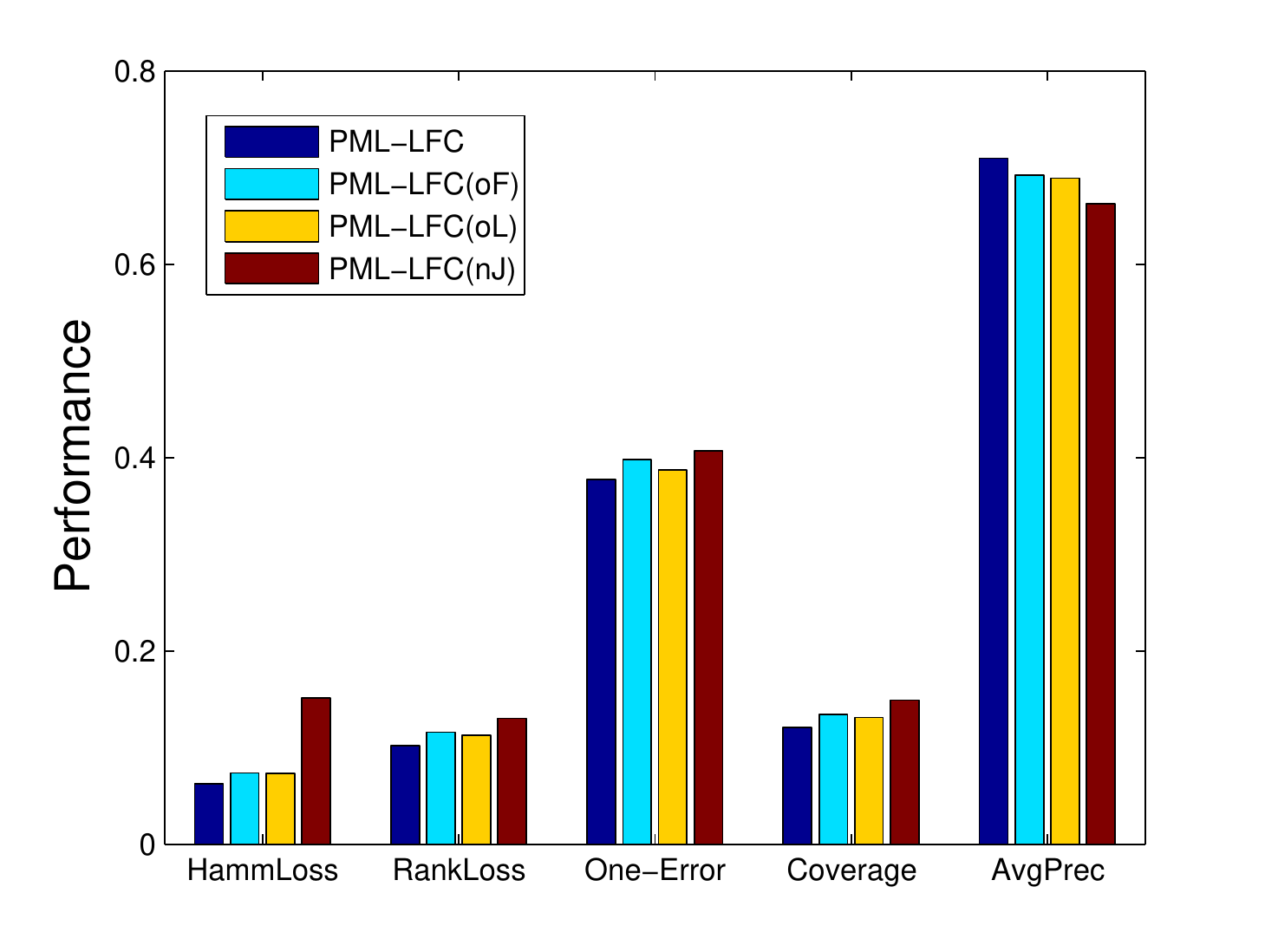}
\caption{The performance of PML-LFC and its degenerated variants on the \emph{slashdot} dataset. For the first four evaluation metrics, the lower the value, the better the performance is. For AvgPrec, the higher the value, the better the performance is.}
\label{fig1}
 \label{figure3}
\end{figure}

We perform ablation study to further study the effectiveness of PML-LFC. For this purpose, we introduce  three variants of PML-LFC, namely, PML-LMF(oF), PML-LFC(oL) and PML-LFC(nJ). PML-LFC(oF) only uses feature similarity, PML-LFC(oL) only utilizes the semantic similarity. PML-LFC(nJ) does not jointly optimize the latent label matrix and the predictor in a unified objective function, it firstly optimizes the latent label matrix and then the multi-label predictor. Fig. \ref{figure3} shows the results of these  variants and PML-LFC on the \emph{slashdot} dataset. All the experimental settings are the same as previous section.

From Fig. \ref{figure3}, we can find that PML-LFC has the lowest HammLoss, RankLoss, One-Error, Coverage, and the highest AvgPrec among the four comparing methods. Neither the feature similarity nor the semantic similarity alone induces a comparable multi-label predictor with PML-LFC. In addition, PML-LFC(oF) and PML-LFC(oL) have similar performance with each other, which indicate that both the feature and label information can be used to induce a multi-label predictor.  PML-LFC leverages both the label and feature information, it induces a less error-prone multi-label classifier and achieves a better classification performance than these two variants. PML-LFC(nJ) has the lowest performance across the five evaluation metrics, which corroborates the disadvantage of isolating the confident matrix learning and multi-label predictor training. This study further confirms that both the feature and label information of multi-label data should be appropriately used for effective partial multi-label learning, and our alternative optimization procedure has a reciprocal reinforce effect for the predictor and the latent label matrix.

\begin{figure*}[h!t]
\centering
\subfigure[Performances vs. $\alpha$]{\label{figure4a}\includegraphics[width=5cm, height=3cm]{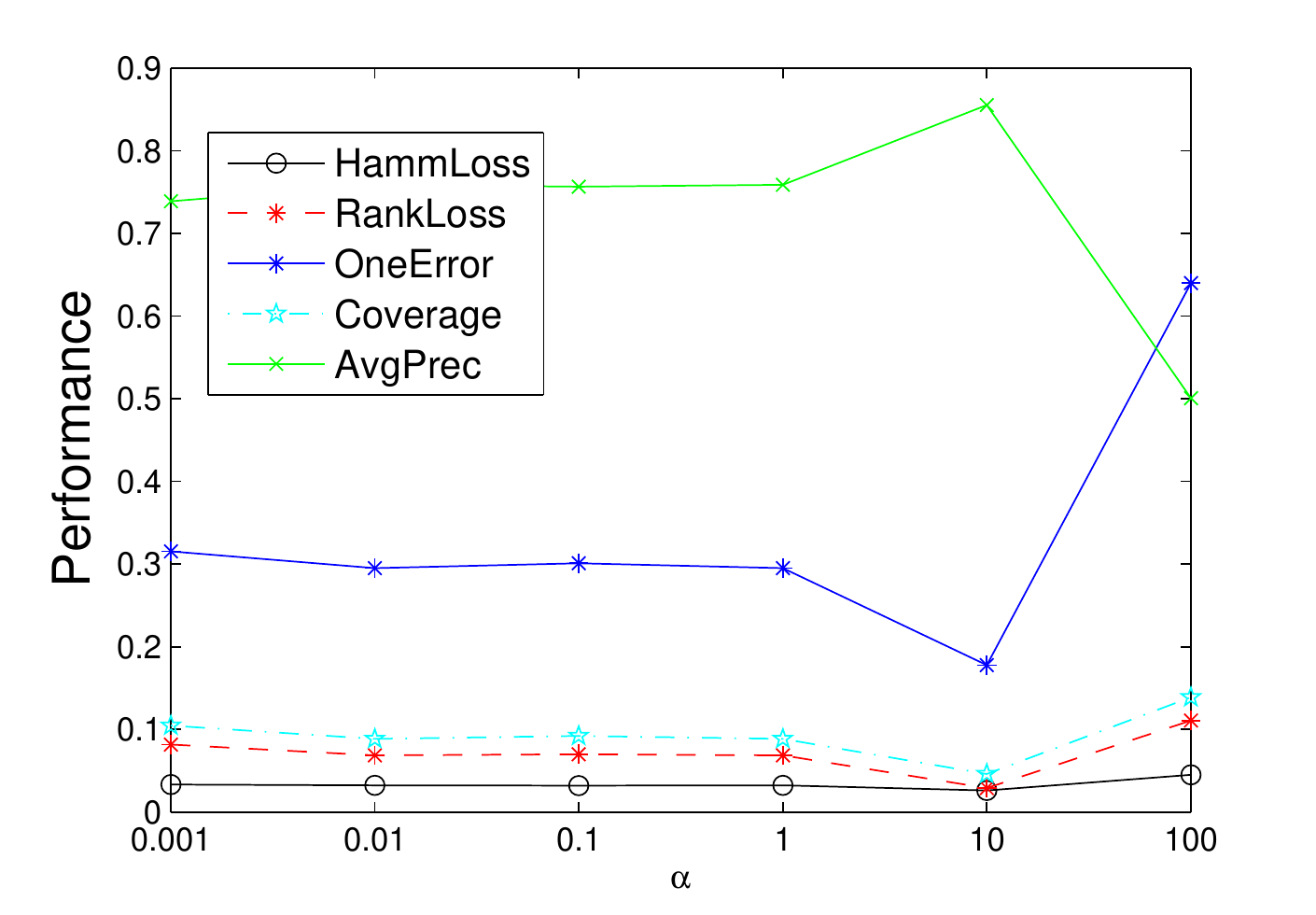}}
\subfigure[Performance vs. $\beta$]{\label{figure4b}\includegraphics[width=5cm, height=3cm]{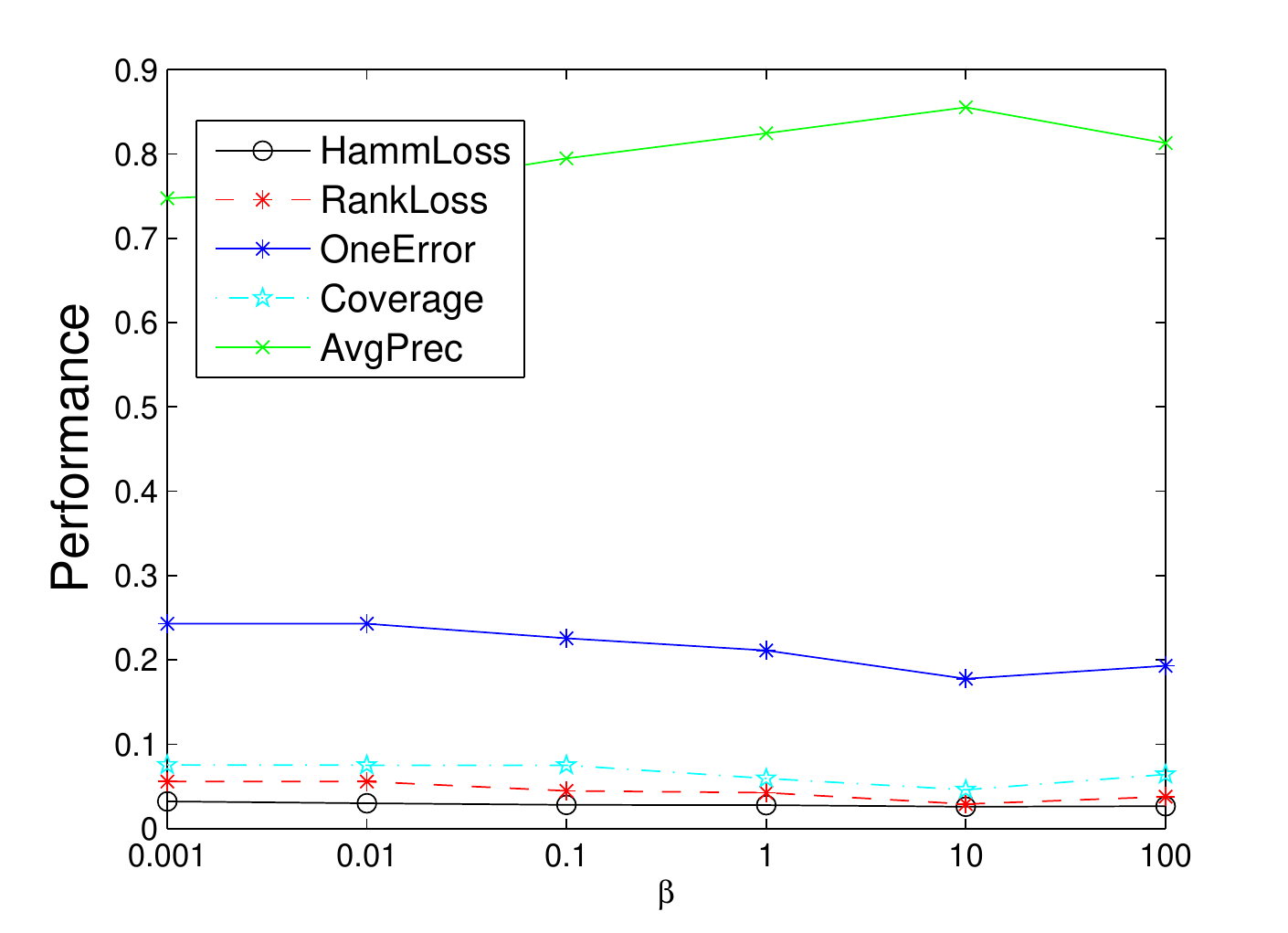}}
\caption{Results of PML-LFC under different input values of $\alpha$ and $\beta$.}
\label{figure4}
\end{figure*}

To investigate the sensitivity of $\alpha$ and $\beta$, we vary $\alpha$ and $\beta$ in the range of \{0.001, 0.01, 0.1, 1, 10, 100\} for PML-LFC on the medical dataset. The experimental results (measured by the five evaluation metrics) are shown in Fig. \ref{figure4}. The results on other datasets give similar observations.  From Fig. \ref{figure4}(a), we can observe that, when $\alpha=10$, PML-LFC achieves the best performance. This observation suggests that it's necessary to consider the low-rank label correlation for partial multi-label learning. When $\alpha$ is too large or too small, the label correlation is underweighted or overweighted, thus the performance manifests a reduce. From Fig. \ref{figure4}(b), we can see that PML-LFC achieves the best performance when $\beta =10$. When $\beta$ is too small, the feature similarity and semantic similarity of multi-label instances are not well accounted, which leads to a poor performance. When $\beta$ is too large (i.e., 100), PML-LFC also achieves a poor performance, as it excessively overweights the feature similarity and semantic similarity, but underweights the prediction model. From this analysis, we adopt $\alpha=10$ and $\beta=10$ for experiments.

\section{Conclusions}
\label{sec:concl}
We investigated the partial multi-label learning problem and proposed an approach called PML-LFC, which leverages the feature and label information for effective multi-label classification. PML-LFC takes into account the negative information between labels and features of partial multi-label data. Extensive experiments on PML datasets from different domains demonstrate the effectiveness of PML-LFC. We are planning to incorporate the abundant unlabeled data for effective extreme partial multi-label learning with a large label space.
\bibliographystyle{splncs04}
\bibliography{ref}

\end{document}